\documentclass[conference]{IEEEtran}
\IEEEoverridecommandlockouts
\usepackage{cite}
\usepackage{amsmath,amssymb,amsfonts}
\usepackage{algorithmic}
\usepackage{graphicx}
\usepackage{textcomp}
\usepackage{xcolor}
\usepackage{booktabs}
\usepackage{multirow}

\def\BibTeX{{\rm B\kern-.05em{\sc i\kern-.025em b}\kern-.08em
    T\kern-.1667em\lower.7ex\hbox{E}\kern-.125emX}}
\begin{document}

\title{QCaption: Video Captioning and Q\&A through Fusion of Large Multimodal Models \\
}

\author{
\IEEEauthorblockN{Jiale Wang\textsuperscript{1}, 
Gee Wah Ng\textsuperscript{1}, 
Lee Onn Mak\textsuperscript{1}, 
Randall Cher\textsuperscript{2}, 
Ng Ding Hei Ryan\textsuperscript{2}, 
Davis Wang\textsuperscript{2}}

\IEEEauthorblockA{\vspace{0.5em}}

\IEEEauthorblockA{\textsuperscript{1}Q Team, Home Team Science and Technology Agency, Singapore\\
\{wang\_jiale, ng\_gee\_wah, mak\_lee\_onn\}@htx.gov.sg}

\IEEEauthorblockA{\textsuperscript{2}Department of Computer Science and Engineering, Nanyang Technological University, Singapore\\
\{rand0017, rng039, dwang023\}@e.ntu.edu.sg}
}

\maketitle

\begin{abstract}
This paper introduces QCaption, a novel video captioning and Q\&A pipeline that enhances video analytics by fusing three models: key frame extraction, a Large Multimodal Model (LMM) for image-text analysis, and a Large Language Model (LLM) for text analysis. This approach enables integrated analysis of text, images, and video, achieving  performance improvements over existing video captioning and Q\&A models; all while remaining fully self-contained, adept for on-premises deployment. Experimental results using QCaption demonstrated up to 44.2\% and 48.9\% improvements in video captioning and Q\&A tasks, respectively. Ablation studies were also performed to assess the role of LLM on the fusion on the results. Moreover, the paper proposes and evaluates additional video captioning approaches, benchmarking them against QCaption and existing methodologies. QCaption demonstrate the potential of adopting a model fusion approach in advancing video analytics.
\end{abstract}


\section{Introduction}

\subsection{Background}

Large Large Language Models (LLM) such as GPT-3 \cite{gpt3_paper}, PaLM \cite{palm_paper}, and LLaMA 2 \cite{llama2_paper},  have captured the world's attention for its remarkable proficiency in understanding and generating human-like text, transforming how we interact with and generate information. 

Recent advancements \cite{clip_paper} in integrating images and text within a unified latent space have catalysed the development of Large Multimodal Models (LMM) such as GPT-4(Vision) \cite{gpt4_paper}, Gemini \cite{gemini_paper}, and LLaVA \cite{llava_paper}. These models possess the capability to interpret images in addition to text, marking a significant leap beyond purely linguistic models, moving AI closer to emulating human intelligence that perceives the world through both visual and textual dimensions. These opened up a whole new dimension of use cases, from image captioning and search, to asking questions about an image.

Despite their multi-modality, most LMMs are primarily designed for interpreting images and text \cite{llava_paper, gemini_paper, gpt4_paper}, lacking the capability to directly process videos. Video analytics, a crucial research domain with extensive applications, holds particular significance for applications like that in homeland security, where critical evidence often exists in video formats such as CCTV footage, body-cam recordings, and phone video recordings of incidents. Developing a versatile AI model and pipeline capable of captioning, analysing, and executing downstream tasks like report writing for videos --- akin to current LMMs' functions with images --- could significantly unlock new automation possibilities, significantly boosting productivity in domains such as homeland security, indexing multimedia databases for search \cite{Abdar2023} or building video recommendation engines.

\subsection{Key contributions}

In this paper, we present QCaption, a comprehensive video captioning and Q\&A pipeline that surpasses existing LMM approaches for video analysis, yielding 44.2\% and 48.9\% improvements in video captioning and Q\&A tasks, respectively. QCaption operates entirely on-premises without the need for external APIs. By conceptualising video captioning as a task of multi-image captioning and caption aggregation, QCaption fuses three models --- a key frame extraction model, an LMM for combined image and text interpretation, and an LLM dedicated to text analysis --- into a video analytics pipeline. This approach allows for the simultaneous analysis of text, images, and video content. Our system builds on the works of LLaVA \cite{llava_paper} and Vicuna \cite{vicuna_paper}, enhancing their capabilities for video content. We demonstrate that QCaption outperforms existing video LMMs, including Video-LLaVA \cite{videollava_paper}, Video-ChatGPT \cite{maaz2023videochatgpt}, and Video-LLaMA \cite{zhang2023videollama}, for video captioning and Q\&A.

Our key contributions include:

\begin{itemize}
    \item The development of QCaption, a novel video captioning pipeline that enables the integrated analysis of text, images, and videos within a single pipeline, unlike previous LMMs that are limited to just image-text or image-video analysis only. Its modular design also allows the frame extraction model, LMM, and LLM to be swapped out as needed.
    \item Through our experiments, we showcase that fusing three models, each adept at different tasks, provides an effective method for video analysis, showing improvements of up to 44.2\% and 48.9\% in video captioning and Q\&A tasks, respectively, over our baselines.
    \item We conduct ablation studies to evaluate the impact of each component within the fusion pipeline across various tasks.
    \item We also proposed and evaluated other video captioning methods, by adapting different key frame sampling approaches and by fusing LLaVA LMM and Vicuna LLM in various ways, and evaluated how they perform against baselines and QCaption.
\end{itemize}

\section{Related Works}

\subsection{Pre-Large Large Language Models (LLM) era}

Traditional Video Captioning and Q\&A models employ Convolutional Neural Networks (CNN) to extract visual features from the video, then returning them as standalone classes or incorporating them into predefined sentence templates to generate a caption or answer a question \cite{Abdar2023, Yousif2023, Guadarrama2013}. Later methods also employ CNNs as the encoder and Recurrent Neural Networks (RNN) for the decoding phase (i.e., language generation) \cite{Alkalouti2021, Yadav2021}. However, the range of requests that can be accepted is highly limited because the model cannot comprehend complex natural language prompts, with use cases confined to just captioning and simple feature based Q\&A. 

\subsection{LLM}

LLMs, such as GPT-3 \cite{gpt3_paper}, have transformed NLP by enabling AI to comprehend and generate complex text. Utilising the transformer architecture, which employs self-attention mechanisms to prioritise words in a sentence to generate contextually relevant text \cite{attention_paper}. Their effectiveness stems from extensive pre-training on large datasets and subsequent fine-tuning for specific applications, such as instruction-tuning for chatbots apt at emulating conversations between humans. Examples include LlaMA 2 \cite{llama2_paper}, Vicuna \cite{vicuna_paper}, and Mistral \cite{Jiang2023}. These LLMs also form the backbone of LMMs, setting them apart from earlier Computer Vision and Video Captioning models.

\subsection{Image Large Multimodal Models (LMMs)}

LMMs build upon LLMs to integrate image processing capabilities with advanced language understanding. There are two main architecture paradigms \cite{videollava_paper}: using LLMs as schedulers, coordinating visual and text models for task-specific applications without end-to-end training, such as HuggingGPT \cite{Shen2023}; and LLMs decoders, aligning image and text data through a two-stage process involving initial auto-regressive pretraining and subsequent refinement with human instruction datasets, such as LLaVA \cite{llava_paper}, mPLUG-Owl \cite{Ye2023}, and Kosmos-2 \cite{Peng2023}. This methodology enables image LMMs to effectively integrate and reason with multimodal data, producing content that aligns with human instructions.

\subsection{Video LMMs}

Building upon image LMMs, video LMMs like Video-LLaVA \cite{videollava_paper}, Video-ChatGPT \cite{maaz2023videochatgpt}, Video-LLaMA \cite{zhang2023videollama} have been developed to handle complex video content, understanding prompts and generating coherent answers beyond simple labels, all of which allows a chatbot-like interaction to analyse videos. However, these models often work best on short clips of a few seconds to 1-2 minutes long and on videos with a single scene, facing challenges with multi-scene videos and capturing temporal relationships \cite{videollava_paper}. 

\subsection{Fusion Techniques}

Information fusion is a techinque for integrating data from multiple sensors or different modalities into a unified representation, often with the goal of enhancing the system's performance \cite{li2023deep}. This technique is primarily divided into two categories: late fusion and early fusion \cite{fusion20231}. Late fusion processes each modality independently and combines them just before the decision-making stage, whereas early fusion merges all modalities at the initial stages, often transforming them into a single latent space \ \cite{fusion20231}. Although not always explicitly discussed, fusion techniques have been utilised in fields such as image/video captioning and question answering.

LLaVA \cite{llava_paper} employs an early fusion method by integrating visual features \( Z_v \) --- extracted from an input image \( X_v \) using the CLIP visual encoder \cite{clip_paper} --- with language embeddings through a projection matrix \( W \). This maps the visual features \( Z_v \) to language tokens \( H_v = W \cdot Z_v \). However, this approach is primarily limited to static images and requires significant adaptations and retraining to handle other information modalities, such as temporal information in videos. VideoLLaVA \cite{videollava_paper} adopts an early fusion approach as well, integrating images and videos into a common feature space. This enables the LLM to leverage a unified visual representation for learning. Similarly, early fusion strategies have been implemented in Video-ChatGPT \cite{maaz2023videochatgpt} and Video-LLaMA \cite{zhang2023videollama} to analyse video content with LMMs.

Most video captioning and Q\&A models have relied on a single-model, early fusion method, which often restricts their efficacy to shorter video clips and overlooks the nuances in longer videos. QCaption introduces a late fusion technique by deploying a pipeline of multiple pre-trained models, each specialising in different modalities. This approach enhances video analytics performance and enables the extraction of richer information from videos of varied lengths.

\section{Methodology}

\subsection{QCaption Fusion Architecture}
\label{qcaption_arch}

\begin{figure*}[ht!]
  \centering
  \includegraphics[width=\textwidth]{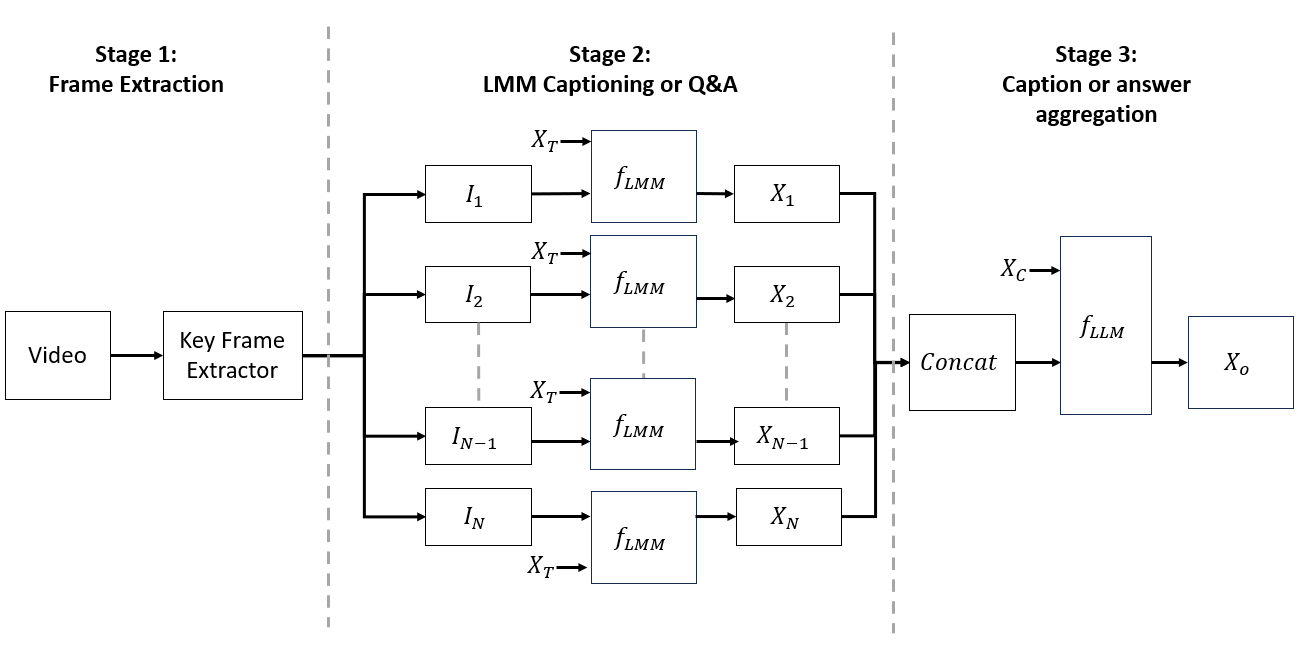}
  \caption{QCaption multi-modal fusion pipeline.}
  \label{fig:qcaption_arch}
\end{figure*}

The QCaption fusion pipeline fuses three models (Fig. \ref{fig:qcaption_arch}) Video-to-image keyframe extractor, 2) Image and text to text captioning and Q\&A LMM, and 3) Text-to-text LLM. 

In stage 1, given input video, a key frame extractor extracts frames from the video as a set of images, \(\mathcal{S} = \{I_1, I_2, \ldots, I_{N-1}, I_N\}\). 

In stage 2, each frame, \(I_i, \text{for } i \in \{1, 2, \ldots, N\}\) is then passed through an LMM model, \(f_{LMM}\), along with the input caption \(X_T\), which is kept consistent for each frame, as denoted by equation \ref{eq3}. In other words, the input prompt posed to the video is repeated across all frames to caption them or generate an answer. 

\begin{equation}
X_i = f_{\text{LMM}}(I_i, X_T), \quad \text{for } i \in \{1, 2, \ldots, N\}
\label{eq3}
\end{equation}

Finally in stage 3, the captions are aggregated to generate a coherent and succinct answer that retains key details. Each individual captions are first combined through a simple string concatenation, per equation \ref{eq4}. Indexes are added in front of each caption prior to concatenation to denote their temporal position (i.e., earliest frame is denoted "0", next "1", and so on).

\begin{equation}
\mathcal{C} = X_1 \Vert X_2 \Vert \ldots \Vert X_N
\label{eq4}
\end{equation}

The concatenated block of captions, \(\mathcal{C}\), is then passed into a LLM model, \(f_{LLM}\), with a prompt \(X_C\), specially designed to aggregate the captions together (section \ref{caption_agg}). From experiments, the recommended prompt to use in this stage varies depending on the task and dataset. This produces the final caption or answer for the video, \(X_o = f_{\text{LLM}}(\mathcal{C}, X_C)\).

For all experiments below, we use \(N=8\), following the frame sampling procedure in Video-LLaVA training stage \cite{videollava_paper}. 

\subsection{Key frame extraction}
\label{katna_sect}

We employed three approaches to frame sampling to assess how it will affect performance depending on dataset used. 

One approach is using the Katna \cite{katna_paper} model for key frame extraction. Katna employs a selective approach in frame selection. Using its frame extractor module, it compares all the video frames that are sufficiently different from previous ones using absolute differences in LUV colorspace, as described in the Katna documentation. Moreover, Katna implements K-Means Clustering of frames using image histograms, enabling a structured organization of frames for enhanced analysis. The selection of the best frame from these clusters is determined based on the variance of Laplacian, facilitating image blur detection to ensure optimal frame representation.  

As an alternative to Katna, we also applied a regular sampling method. Proposed by Gemini \cite{gemini_paper}, frames were sampled at regular intervals under this approach. Each video was divided into equal segments and a frame was extracted from the midpoint of each segment for analysis. 

A third approach involved random sampling, where frames are selected randomly from throughout the video, providing an randomised, varied set of frames for analysis. 

\subsection{Frame Captioning}

Frame captioning involves generating text descriptions for individual frames within a video. This process turns visual information into textual descriptions, making it easier for computers to understand and interpret video content. We used LlaVA \cite{llava_paper} to generate these captions. These models are trained on datasets comprising images and their corresponding descriptive captions. This training enables them to understand the correlation between visual elements and textual descriptions, where LLaVA specialises in generating captions for static frames. 

\subsection{Caption Aggregation}

\label{caption_agg}

Caption aggregation is achieved through the use of the Vicuna LLM  \cite{vicuna_paper}, which synthesises text descriptions from individual video frames into a coherent summary. This process enhances the understanding of video content by filtering out redundancies and selecting crucial information if different frames yielded different information. This ensures that the final summary is both concise and informative, enhancing understanding for users. Prompt engineering was employed through a trial-and-error approach to determine an optimal prompt for caption aggregation.

\subsection{Advanced Sampling - Multi clips method}

As an alternative to the above frame captioning methods, we also propose the mutli-clips captioning method. The multi-clips method involves an approach to video captioning where we sample four 5-second clips at regular intervals from the video. This approach aims to capture more temporal information, for more efficient content management and comprehensive analysis. Once the video is segmented, each subsection is independently processed through VideoLLaVA \cite{videollava_paper} to generate the relevant captions. Following this, the captions generated from each subsection are then passed into the LLM for caption aggregation.  

\subsection{Ablation Studies and Other methods}
\label{other_methods_method}

\subsubsection{Ablation Studies}

We also performed ablation studies to evaluate the impact of the LLM stage on the fusion pipeline. In some experiments, the LLM was removed and the individual frame captions were joint to gather through naive string concatenation. 

\subsubsection{First N-frames method}

We also implemented a method which returned the first N-frames for analysis. Similar to the previously mentioned techniques, we applied both an LLM and non-LLM approach to the captioning method. 

Finally, various combinations of the above were tested. The goal is to evaluate a range of LMM-based video captioning strategies and whether any of them are more suitable for some datasets.

\section{Results}

\begin{table*}[h!]
\centering
\caption{Performance comparison on video captioning and Q\&A benchmarks, between exisiting work and QCaption.}
\label{tab:results_table}
\begin{tabular}{@{}lllllll@{}}
\toprule
\multirow{3}{*}{S/N} & \multirow{3}{*}{} & \multirow{3}{*}{Method}        & \multicolumn{2}{l}{Video Captioning}                             & \multicolumn{2}{l}{Video Q\&A}            \\ \cmidrule(l){4-7} 
                     &                   &                                & \multirow{2}{*}{YouCook2 (val)} & \multirow{2}{*}{MSR-VTT (val)} & \multicolumn{2}{l}{ActivityNet-QA (test)} \\
                     &                   &                                &                                 &                                & Accuracy              & Score             \\ \cmidrule(r){1-7}
1                    & Baselines        & VideoLLaMA                   & -                               & -                              & 12.4                  & 1.10              \\
2                    &                  & VideoChatGPT                   & -                               & -                              & 35.2                  & 2.70              \\
3                    &                   & Video-LLaVA                    & 19.9                            & 28.6                           & 42.5                  & 3.53              \\ \cmidrule(r){1-7}
4                    & QCaption          & Katna + LLaVA + LLM            & \textbf{28.7}                             & 36.8                           & 61.6                  & 3.74              \\
5                    &                   & Regular Sampling + LLaVA + LLM & 26.3                            & 36.5                           & \textbf{63.3}                  & 3.87                \\ 
6                    &                                & Random Sampling + LLaVA + LLM     & 26.0                            & 35.1                           & 58.2                  & 3.78     \\
7                    &                                & Multiclips: Clip Sampling + Video-LLaVA + LLM       & 25.5                            & \textbf{41.7}                           & 53.1                  & 3.52              \\
\cmidrule(r){1-7}
\end{tabular}
\end{table*}

\subsection{Benchmark Datasets}

To benchmark QCaption and other video captioning and Q\&A approaches against existing work like Video-LLaVA \cite{videollava_paper}, we employed three commonly reported datasets: YouCook2 \cite{youcook2_paper} and MSR-VTT \cite{msrvtt_paper} for video captioning, and ActivityNet-QA \cite{activitynetQA_paper} for video Q\&A.

YouCook2 comprises untrimmed, third-person viewpoint videos from 89 global cooking recipes, each annotated with procedural steps in imperative English sentences. Following the Gemini paper's approach \cite{gemini_paper}, we used the YouCook2 validation set comprising $\approx$450 videos.

MSR-VTT is a comprehensive large-scale video description dataset with a range of web clips extracted from a commercial video search engine, accompanied by an average of 20 associated captions per video, annotated by AMT workers. Topics depicted in MSR-VTT include news reports, animals, transportation, cooking, etc. We used the MSR-VTT validation set comprising $\approx$500 videos.

ActivityNet-QA is a fully annotated, large-scale VideoQA dataset with web videos and on average 10 QA pairs for each video. The videos are derived from the ActivityNet \cite{activitynet_paper} dataset,  encompassing videos on a diverse array of complex human activities relevant to daily life. Likewise, following authors of Gemini \cite{gemini_paper}, we employed the ActivityNet-QA test set comprising 800 videos.

\subsection{Evaluation Metrics}

\subsubsection{CIDEr}
\label{cider_sect}

For the task of video captioning using YouCook2 and MSR-VTT datasets, after captions were generated using the various pipelines in our experiments, we compared them against the ground truth captions (from the dataset) using the \textbf{CIDEr} metric \cite{cider_paper}. This aligns with the metric used by authors of Gemini \cite{gemini_paper}. Originally developed for image captioning but later adopted for video captioning, CIDEr quantifies that for given video \(I_i\), how well its generated caption \(c_i\) aligns with the consensus of a set of ground truth descriptions \(S_i = \{s_{i1}, \ldots, s_{im}\}\). During evaluation, each caption is represented by a set of \textit{n-grams} \(\omega_k\) comprising 1-4 words. The Term Frequency Inverse Document Frequency (TF-IDF) for each \textit{n-gram}, \(g_k(s_{ij})\), is then computed using equation \ref{eq1}, which accounts for how often \textit{n-grams} in the generated caption is present in ground truths, and that \textit{n-grams} not in ground truth should not appear in generated caption. Note that \(h_k(\cdot)\) the number of times a \textit{n-gram} occurs in a caption.

\begin{equation}
g_k(s_{ij}) = \frac{h_k(s_{ij})}{\sum_{\omega_l \in \Omega} h_l(s_{ij})} \log \left( \frac{|I|}{\sum_{I_p \in I} \min(1, \sum_{q} h_k(s_{pq}))} \right),
\label{eq1}
\end{equation}

The CIDEr score for \textit{n-gram} of length \textit{n} is then computed using equation \ref{eq2}, which compares the average cosine similarity between generated and ground truth captions. 

\begin{equation}
\text{CIDEr}_n(c_i, S_i) = \frac{1}{m} \sum_j \frac{g^n(c_i) \cdot g^n(s_{ij})}{\|g^n(c_i)\| \|g^n(s_{ij})\|}
\label{eq2}
\end{equation}
Scores from individual \textit{n-grams} are then summed up to yield the final CIDEr score, per equation \ref{eq3}.

\begin{equation}
\text{CIDEr}(c_i, S_i) = \sum_{n=1}^{N} w_n \text{CIDEr}_n(c_i, S_i),
\label{eq3}
\end{equation}

\subsubsection{Video-ChatGPT metric}
\label{videoGPT_sect}

To evaluate the quality of answers generated for Q\&A tasks using the ActivityNet-QA dataset, we employed the Video-ChatGPT evaluation protocol \cite{videollava_paper}, also used by Video-LLaVA \cite{videollava_paper} and Gemini \cite{gemini_paper}. The metric was designed to evaluate the text generation capabilities of video-based conversational models (e.g., video Q\&A). This evaluation encompasses five critical dimensions: Correctness of Information, Detailed Orientation, Contextual Understanding, Temporal Understanding, and Consistency.  For each input question for a video, every generated and ground truth pair are evaluated in the above dimensions using ChatGPT "gpt-3.5-turbo" version, called through the OpenAI API in the evaluation script.

\subsubsection{Qualitative}
We also performed qualitative experiments using videos of a few minutes long, each featuring different scenes or point-of-views, to compare the prowess of each pipeline in capturing key details and the sequence of events depicted. 

\subsection{Baseline}

To validate the proposed QCaption multi modal fusion pipeline, we compared it against the following notable LMM-based video captioning models:
\begin{itemize}
    \item Video-LLaMA \cite{zhang2023videollama}: LMM that employs a common visual encoder to understand both images and videos.
    \item Video-ChatGPT \cite{maaz2023videochatgpt}: LMMs that assign a unique encoder to each modality, for the ability to understand images or videos via multiple projection layers.
    \item Video-LLaVA \cite{videollava_paper}: LMM that aligns image and video representations before projection, which enables a unified visual representation of both image and video. Employs the Vicuna-v1.5 LLM, ViT-L/14 visual encoder, and LlaMA text tokeniser, just like LLaVA-v1.5 \cite{llava_paper}. Video-LLaVA outperformed both Video-LLaMA and Video-ChatGPT.
\end{itemize}

The baselines all employ an early fusion approach; QCaption utilises late fusion.

QCaption is designed to be LMM and LLM agnostic, one can swap out any part of the pipeline for other models. For clarity of comparison, we built QCaption on top of LLaVA-v1.5 \cite{llava_paper}, a LMM for image captioning and Q\&A that shares ample similarities to the architecture of Video-LLaVA. The LLM used is Vicuna-v1.5 \cite{vicuna_paper}. 

Performance benchmarks for Video-LLaMA and Video-ChatGPT were adapted from the Video-LLaVA paper; results for Video-LLaVA were from experiments closely following the experimental procedure stipulated in the paper \cite{videollava_paper}. For YouCook2 and MSR-VTT captioning datasets, the CIDEr \cite{cider_paper} metric was used (section \ref{cider_sect}); for ActivityNet-QA Q\&A dataset, the Video-ChatGPT evaluation approach (section \ref{videoGPT_sect}) was adopted.

\subsection{QCaption}

Using the QCaption pipeline (section \ref{qcaption_arch}), two frame extraction methods were tested: using Katna key frame sampling (section \ref{katna_sect}) and Regular Sampling (taking frames at equally spaced intervals in the video). 8 frames were sampled for each video. For both approaches, frames were annotated using LLaVA-1.5 LMM and the captions were aggregated using the Vicuna-v1.5 LLM. Both methods yielded improvements over the baseline Video-LLaVA approach for some captioning and Q\&A tasks.

Referring to Table \ref{tab:results_table}, Using Katna + LLaVA + LLM yielded 44.2\%, 28.7\%, 44.9\% improvements for video captioning (YouCook2, MSR-VTT) and Q\&A (ActivityNet-QA) tasks respectively. Using Regular Sampling + LLaVA + LLM yielded improvements of 32.2\%, 27.6\% and 48.9\% respectively on captioning (YouCook2, MSR-VTT) and Q\&A (ActivityNet-QA). 

The results indicate that regular sampling of frames yields better results for Q\&A tasks, but key frame extractions (using Katna) is more useful for captioning. In general, in tasks such as general scene captioning or answering general Q\&As, treating video captioning as a multi-image captioning approach --- by late fusion of frame extraction algorithms, LMM, and LLM --- performs better compared to constructing a video-image-text model from scratch, which was the case of the baselines. Additionally, this strategy greatly reduces the resources needed for data curation and training of a new model. 

\subsection{Advanced Sampling -  Multiclips}

QCaption took this idea a step further by breaking a long video analysis problem into one of analysing multiple small videos. Similar to the image-based pipeline, this approach first samples short clips of 5s long at regular intervals, and then iteratively caption them using Video-LLaVA. The captions are then aggregated using a LLM. Per Table \ref{tab:results_table}, this approach yielded notable 45.8\% and 24.9\% improvements for both video captioning (MSR-VTT) and Q\&A (ActivityNet-QA) tasks. 

\subsection{Ablation Studies and Other methods}
\label{other_methods}
\begin{table*}[h!]
\centering
\caption{Performance comparison on video captioning and Q\&A benchmarks, between QCaption and other possible video captioning techniques.}
\label{tab:results_table_2}
\begin{tabular}{@{}lllllll@{}}
\toprule
\multirow{3}{*}{S/N} & \multirow{3}{*}{}              & \multirow{3}{*}{Method}           & \multicolumn{2}{l}{Video Captioning}                             & \multicolumn{2}{l}{Video Q\&A}            \\ \cmidrule(l){4-7} 
                     &                                &                                   & \multirow{2}{*}{YouCook2 (val)} & \multirow{2}{*}{MSR-VTT (val)} & \multicolumn{2}{l}{ActivityNet-QA (test)} \\
                     &                                &                                   &                                 &                                & Accuracy              & Score             \\ \cmidrule(r){1-7}
1                    & \multirow{3}{*}{QCaption}      & Katna + LLaVA + LLM               & \textbf{28.7}                             & 36.8                           & 61.6                  & 3.74              \\ 
2                   &                                & Regular Sampling + LLaVA + LLM    & 26.3                            & 36.5                           & \textbf{63.3}                  & \textbf{3.87}  \\
3                    &                                & Random Sampling + LLaVA + LLM     & 26.0                            & 35.1                           & 58.2                  & 3.78              \\
4                   &                                & Multiclips: Clip Sampling + Video-LLaVA + LLM       & 25.5                            & 41.7                           & 53.1                  & 3.52              \\ \cmidrule(r){1-7}
5                    &  \multirow{2}{*}{Ablation Studies}                              & Regular Sampling + LLaVA (no LLM) & 21.3                            & \textbf{52.3}                           & 55.9                  & 3.47              \\
6                    &                                & Random Sampling + LLaVA (no LLM)  & 21.1                            & \textbf{52.3}                           & 55.0                  & 3.42              \\ \cmidrule(r){1-7}
7                   & \multirow{3}{*}{Other methods}& First N frames + LLaVA (no LLM)   & 9.4                             & 42.9                           & 45.1                  & 3.34              \\
8                   &                                & Sample 1 frame only + LLaVA       & 13.9                            & 34.1                           & 57.4                  & 3.76              \\ \cmidrule(r){1-7}
\end{tabular}
\end{table*}

We also conducted a comparison of QCaption with other possible video captioning methods and performed ablation studies by disabling or swapping out certain stages of QCaption, all of which is detailed in Section \ref{other_methods_method}. Referencing table \ref{tab:results_table_2}, the best video captioning approach varies depending on the task and type of answers expected in a given dataset.

For the YouCook2 dataset, which predominantly features detailed captions outlining specific ingredients and recipe steps, we discovered that regular sampling of frames, caption generation via LMM, and subsequent integration through simple string concatenation (without LLM aggregation) still outperforms the baseline with a 7.04\% improvement. With the use of the LLM, as proposed by QCaption, the results show a further improvement of 19.0\%.  Conversely, employing the first N frames or single frame sampling method all yields vastly inferior results, owing to the loss of information from lack of samples. The above all supports the effectiveness of the Fusion approach proposed by QCaption.

For the ActivityNet-QA dataset, focusing on Q\&A tasks, the inclusion of LLMs emerged as crucial for generating coherent and precise answers. Our ablation studies demonstrated that eliminating the LLM stage led to a performance decline of up to \(10.3\%\), compared to QCaption's LLM-integrated methods. Interestingly, employing a single-frame sampling strategy yielded outcomes on par with QCaption's top-performing method, suggesting that straightforward questions might be effectively addressed using just one indicative frame from the video. Nonetheless, the combination of QCaption's random sampling with LMM and subsequent LLM processing consistently delivered superior results, underscoring the significance of this fusion approach in video Q\&A.

For the MSR-VTT dataset, eliminating LLM-aggregation led to  improvements, with up to a 43.3\% increase in performance over the LLM-enabled QCaption method.  This likely stems from MSR-VTT datasets employing much shorter, often one-word long descriptions. Using a LLM could potentially abstract away that one-word that was correctly captured during the video captioning process. Interestingly, employing the first N frames (without LLM) achieved results on par with QCaption's top-performing approach (Multiclips). This indicates that, for MSR-VTT videos, capturing the initial scene alone is often adequate for a comprehensive understanding of the video, likely due to the absence of significant scene transitions, and one-word answer. The case for MSR-VTT is rather unique, as general video captioning / Q\&A  tasks like YouCook2 and ActivityNet-QA require answers in complete sentences, which could be effectively captured using the QCaption pipeline.

In all, QCaption's fusion pipeline consistently outperforms baselines like Video-LLaVA for both captioning and Q\&A datasets. Notably, for both tasks where a comprehensive, complete answer is expected, the LLM is often an important final step to aggregate a coherent answer.

\subsection{Qualitative Experiments}

\begin{figure*}[ht!]
  \centering
  \includegraphics[width=\textwidth]{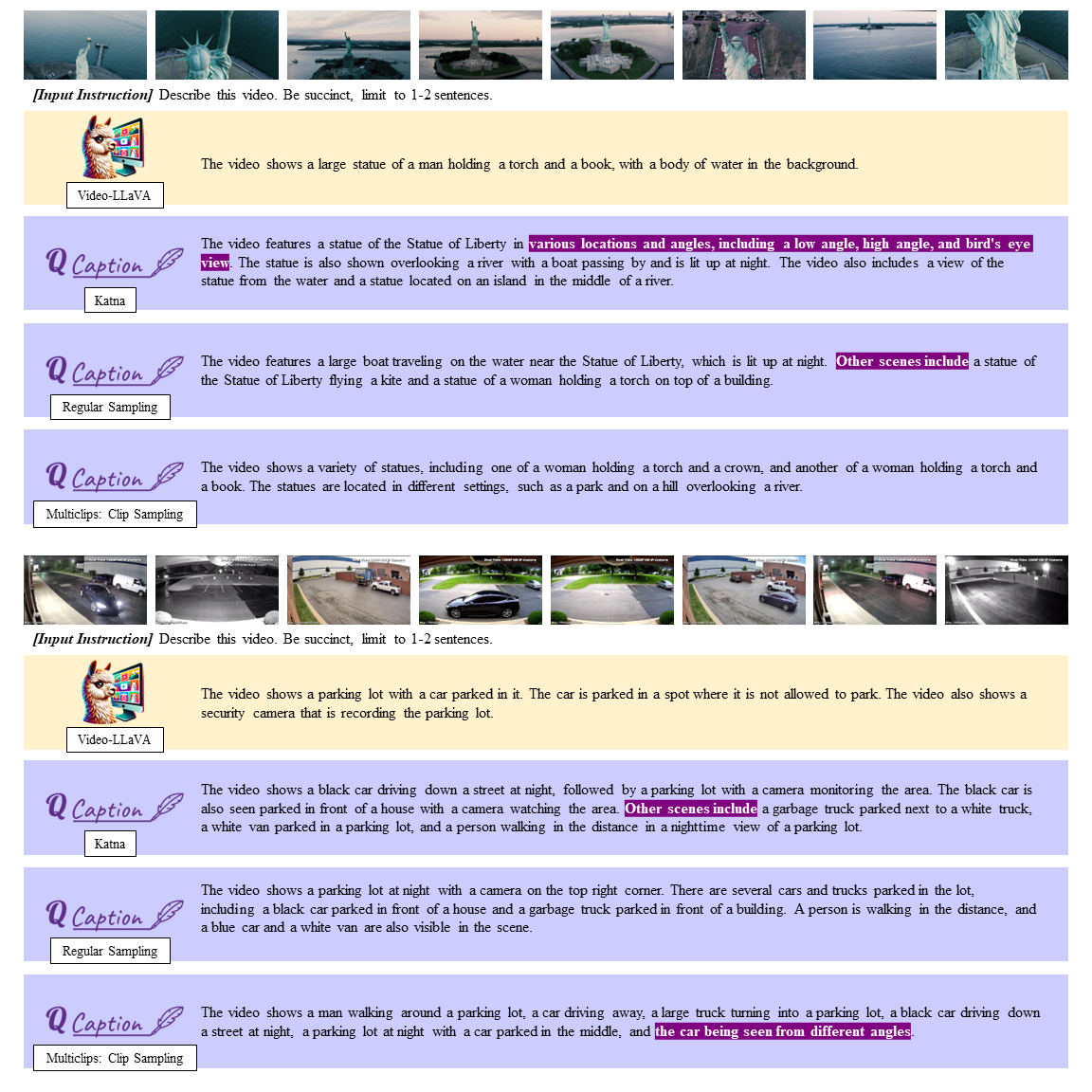}
  \caption{Comparison of Video-LLaVA and QCaption's video captioning capabilities}
  \label{fig:qualitative_comparison}
\end{figure*}

In Fig. \ref{fig:qualitative_comparison}, we curated two illustrative examples to meticulously assess the video captioning prowess of QCaption. We systematically juxtapose its performance against Video-LLaVA using a pair of carefully selected videos. These videos are deliberately chosen to encompass diverse camera angles, thereby elevating the intricacy of the video inputs. The primary objective is to gauge the nuanced capabilities of each model in the realm of video captioning.

Our overarching observation indicates that both QCaption and Video-LLaVA adequately interpret and articulate the content of the videos in accordance with the provided instructions. However, QCaption demonstrates a enhanced proficiency in capturing crucial information spanning multiple scenes within each video. Its prowess becomes evident in discerning changes in camera angles and perspectives, particularly highlighted in \textcolor{purple}{\textbf{purple}} in the presented examples. This makes QCaption's fusion approach preferrable for understanding longer videos with multiple scene changes.

\section{Conclusion and Future Work}

In this paper, we introduced QCaption, a video captioning and Q\&A pipeline that stands out for its fusion of three distinct models: a key frame extraction model, an LMM for image-text analysis, and an LLM for text analysis. This integrated approach enables simultaneous analysis of text, images, and video, achieving superior performance both quantitative and qualitative experiments for video captioning and Q\&A tasks, all while using pre-trained models. We also performed ablation studies to examine how certain components of the pipeline (or the lack thereof) can impact performance of different tasks, and proposed and bench-marked other possible video captioning pipelines.

\textbf{Future work} We observed that outcome of the caption aggregation stage is highly dependent on the input prompt for the LLM, for it determines what information is retained and abstracted away. We prompt engineered using a trial and error approach to qualitatively determine a prompt that yields decent results. There is room to dive deeper into how the aggregation prompt affects quantitative results, and whether the prompt to use varies between datasets. There is also room to test on more recent image LMM models like Gemini Pro \cite{gemini_paper} and GPT-4V \cite{gpt4_paper} to further improve the performance while retaining the same pipeline. Finally, the modular approach to fusion also affords potential to incorporate other modalities, such as speech to text models into the pipeline, which further enrich the information that the LLM can work with, to produce more comprehensive captions and answers. QCaption can serve as a baseline for fusion-based approaches to video captioning that do not require costly architecture modification and retraining of models.

\section*{Acknowledgment}

This work is funded by the Home Team Science and Technology Agency (HTX), a statutory board under the Ministry of Home Affairs (MHA), Singapore.


\bibliographystyle{unsrt}
\bibliography{bibtex.bib}

@misc{li2023deep,
      title={Deep Model Fusion: A Survey}, 
      author={Weishi Li and Yong Peng and Miao Zhang and Liang Ding and Han Hu and Li Shen},
      year={2023},
      eprint={2309.15698},
      archivePrefix={arXiv},
      primaryClass={cs.LG}
}

@article{fusion20231,
   author = {Maciej Pawłowski and Anna Wróblewska and Sylwia Sysko-Romańczuk},
   doi = {10.3390/S23052381},
   issn = {14248220},
   issue = {5},
   journal = {Sensors (Basel, Switzerland)},
   month = {3},
   pmid = {36904585},
   publisher = {Multidisciplinary Digital Publishing Institute  (MDPI)},
   title = {Effective Techniques for Multimodal Data Fusion: A Comparative Analysis},
   volume = {23},
   url = {/pmc/articles/PMC10007548/ /pmc/articles/PMC10007548/?report=abstract https://www.ncbi.nlm.nih.gov/pmc/articles/PMC10007548/},
   year = {2023},
}

@article{palm_paper,
   author = {Aakanksha Chowdhery and Sharan Narang and Jacob Devlin},
   month = {4},
   title = {PaLM: Scaling Language Modeling with Pathways},
   url = {https://arxiv.org/abs/2204.02311v5},
   year = {2022},
}

@article{gpt3_paper,
   author = {Tom B. Brown and Benjamin Mann and Nick Ryde},
   issn = {10495258},
   journal = {Advances in Neural Information Processing Systems},
   month = {5},
   publisher = {Neural information processing systems foundation},
   title = {Language Models are Few-Shot Learners},
   volume = {2020-December},
   url = {https://arxiv.org/abs/2005.14165v4},
   year = {2020},
}

@article{llama2_paper,
   author = {Hugo Touvron and Louis Martin and Kevin Stone},
   month = {7},
   title = {Llama 2: Open Foundation and Fine-Tuned Chat Models},
   url = {https://arxiv.org/abs/2307.09288v2},
   year = {2023},
}

@article{gpt4_paper,
   author = {OpenAI},
   month = {3},
   title = {GPT-4 Technical Report},
   url = {http://arxiv.org/abs/2303.08774},
   year = {2023},
}

@article{gemini_paper,
   author = {Gemini Team},
   month = {12},
   title = {Gemini: A Family of Highly Capable Multimodal Models},
   url = {http://arxiv.org/abs/2312.11805},
   year = {2023},
}

@article{llava_paper,
   author = {Haotian Liu and Chunyuan Li and Qingyang Wu and Yong Jae Lee},
   month = {4},
   title = {Visual Instruction Tuning},
   url = {https://arxiv.org/abs/2304.08485v2},
   year = {2023},
}

@article{clip_paper,
   author = {Alec Radford and Jong Wook Kim and Chris Hallacy and Aditya Ramesh and Gabriel Goh and Sandhini Agarwal and Girish Sastry and Amanda Askell and Pamela Mishkin and Jack Clark and Gretchen Krueger and Ilya Sutskever},
   isbn = {9781713845065},
   issn = {26403498},
   journal = {Proceedings of Machine Learning Research},
   month = {2},
   pages = {8748-8763},
   publisher = {ML Research Press},
   title = {Learning Transferable Visual Models From Natural Language Supervision},
   volume = {139},
   url = {https://arxiv.org/abs/2103.00020v1},
   year = {2021},
}

@article{Abdar2023,
   author = {Moloud Abdar and Meenakshi Kollati and Swaraja Kuraparthi and Farhad Pourpanah and Daniel McDuff and Mohammad Ghavamzadeh and Shuicheng Yan and Abduallah Mohamed and Abbas Khosravi and Erik Cambria and Fatih Porikli},
   month = {4},
   title = {A Review of Deep Learning for Video Captioning},
   url = {https://arxiv.org/abs/2304.11431v1},
   year = {2023},
}

@article{videollava_paper,
   author = {Bin Lin and Yang Ye and Bin Zhu and Jiaxi Cui and Munang Ning and Peng Jin and Li Yuan},
   month = {11},
   title = {Video-LLaVA: Learning United Visual Representation by Alignment Before Projection},
   url = {https://arxiv.org/abs/2311.10122v2},
   year = {2023},
}

@article{Yousif2023,
   author = {Adel Jalal Yousif and Mohammed H. Al-Jammas},
   doi = {10.1016/J.PRIME.2023.100372},
   issn = {2772-6711},
   journal = {e-Prime - Advances in Electrical Engineering, Electronics and Energy},
   month = {12},
   pages = {100372},
   publisher = {Elsevier},
   title = {Exploring deep learning approaches for video captioning: A comprehensive review},
   volume = {6},
   year = {2023},
}

@article{Guadarrama2013,
   author = {Sergio Guadarrama and Niveda Krishnamoorthy and Girish Malkarnenkar and Subhashini Venugopalan and Raymond Mooney and Trevor Darrell and Kate Saenko},
   doi = {10.1109/ICCV.2013.337},
   isbn = {9781479928392},
   journal = {Proceedings of the IEEE International Conference on Computer Vision},
   pages = {2712-2719},
   publisher = {Institute of Electrical and Electronics Engineers Inc.},
   title = {Youtube2text: Recognizing and describing arbitrary activities using semantic hierarchies and zero-shot recognition},
   year = {2013},
}

@article{Alkalouti2021,
   author = {Hanan Nasser Alkalouti and Mayada Ahmed Al Masre},
   doi = {10.1109/IEMTRONICS52119.2021.9422600},
   isbn = {9781665440677},
   journal = {2021 IEEE International IOT, Electronics and Mechatronics Conference, IEMTRONICS 2021 - Proceedings},
   month = {4},
   publisher = {Institute of Electrical and Electronics Engineers Inc.},
   title = {Encoder-decoder model for automatic video captioning using yolo algorithm},
   year = {2021},
}

@article{Yadav2021,
   author = {Naveen Yadav and Dinesh Naik},
   doi = {10.1109/I2CT51068.2021.9417907},
   isbn = {9781728188768},
   journal = {2021 6th International Conference for Convergence in Technology, I2CT 2021},
   month = {4},
   publisher = {Institute of Electrical and Electronics Engineers Inc.},
   title = {Generating Short Video Description using Deep-LSTM and Attention Mechanism},
   year = {2021},
}

@article{attention_paper,
   author = {Ashish Vaswani and Noam Shazeer and Niki Parmar and Jakob Uszkoreit and Llion Jones and Aidan N. Gomez and Łukasz Kaiser and Illia Polosukhin},
   isbn = {1706.03762v7},
   issn = {10495258},
   journal = {Advances in Neural Information Processing Systems},
   month = {6},
   pages = {5999-6009},
   publisher = {Neural information processing systems foundation},
   title = {Attention Is All You Need},
   volume = {2017-December},
   url = {https://arxiv.org/abs/1706.03762v7},
   year = {2017},
}

@article{vicuna_paper,
   author = {Baolin Peng and Chunyuan Li and Pengcheng He and Michel Galley and Jianfeng Gao},
   month = {4},
   title = {Instruction Tuning with GPT-4},
   url = {http://arxiv.org/abs/2304.03277},
   year = {2023},
}

@article{Jiang2023,
   author = {Albert Q. Jiang and Alexandre Sablayrolles and Arthur Mensch and Chris Bamford and Devendra Singh Chaplot and Diego de las Casas and Florian Bressand and Gianna Lengyel and Guillaume Lample and Lucile Saulnier and Lélio Renard Lavaud and Marie-Anne Lachaux and Pierre Stock and Teven Le Scao and Thibaut Lavril and Thomas Wang and Timothée Lacroix and William El Sayed},
   month = {10},
   title = {Mistral 7B},
   url = {https://arxiv.org/abs/2310.06825v1},
   year = {2023},
}

@article{youcook2_paper,
   author = {Luowei Zhou and Chenliang Xu and Jason J. Corso},
   doi = {10.1609/aaai.v32i1.12342},
   isbn = {9781577358008},
   issn = {2159-5399},
   journal = {32nd AAAI Conference on Artificial Intelligence, AAAI 2018},
   month = {3},
   pages = {7590-7598},
   publisher = {AAAI press},
   title = {Towards Automatic Learning of Procedures from Web Instructional Videos},
   url = {https://arxiv.org/abs/1703.09788v3},
   year = {2017},
}

@misc{msrvtt_paper,
   author = {Jun Xu and Tao Mei and Ting Yao and Yong Rui},
   month = {6},
   title = {MSR-VTT: A Large Video Description Dataset for Bridging Video and Language},
   url = {https://www.microsoft.com/en-us/research/publication/msr-vtt-a-large-video-description-dataset-for-bridging-video-and-language/},
   year = {2016},
}

@article{activitynetQA_paper,
   author = {Zhou Yu and Dejing Xu and Jun Yu and Ting Yu and Zhou Zhao and Yueting Zhuang and Dacheng Tao},
   doi = {10.1609/aaai.v33i01.33019127},
   isbn = {9781577358091},
   issn = {2159-5399},
   journal = {33rd AAAI Conference on Artificial Intelligence, AAAI 2019, 31st Innovative Applications of Artificial Intelligence Conference, IAAI 2019 and the 9th AAAI Symposium on Educational Advances in Artificial Intelligence, EAAI 2019},
   month = {6},
   pages = {9127-9134},
   publisher = {AAAI Press},
   title = {ActivityNet-QA: A Dataset for Understanding Complex Web Videos via Question Answering},
   url = {https://arxiv.org/abs/1906.02467v1},
   year = {2019},
}

@inproceedings{activitynet_paper,
  title={ActivityNet: A Large-Scale Video Benchmark for Human Activity Understanding},
  author={Fabian Caba Heilbron, Victor Escorcia, Bernard Ghanem and Juan Carlos Niebles},
  booktitle={Proceedings of the IEEE Conference on Computer Vision and Pattern Recognition},
  pages={961--970},
  year={2015}
}

@article{cider_paper,
   author = {Ramakrishna Vedantam and C. Lawrence Zitnick and Devi Parikh},
   doi = {10.1109/CVPR.2015.7299087},
   isbn = {9781467369640},
   issn = {10636919},
   journal = {Proceedings of the IEEE Computer Society Conference on Computer Vision and Pattern Recognition},
   month = {11},
   pages = {4566-4575},
   publisher = {IEEE Computer Society},
   title = {CIDEr: Consensus-based Image Description Evaluation},
   volume = {07-12-June-2015},
   url = {https://arxiv.org/abs/1411.5726v2},
   year = {2014},
}

@misc{katna_paper,
   author = {Mayank Jain and Nitin Katyal and Alok},
   title = {GitHub - keplerlab/katna: Tool for automating common video key-frame extraction, video compression and Image Auto-crop/Image-resize tasks},
   url = {https://github.com/keplerlab/katna},
   year = {2021},
}

@article{Shen2023,
   author = {Yongliang Shen and Kaitao Song and Xu Tan and Dongsheng Li and Weiming Lu and Yueting Zhuang and Zhejiang University and Microsoft Research Asia},
   month = {3},
   title = {HuggingGPT: Solving AI Tasks with ChatGPT and its Friends in Hugging Face},
   url = {https://arxiv.org/abs/2303.17580v4},
   year = {2023},
}

@article{Ye2023,
   author = {Qinghao Ye and Haiyang Xu and Guohai Xu and Jiabo Ye and Ming Yan and Yiyang Zhou and Junyang Wang and Anwen Hu and Pengcheng Shi and Yaya Shi and Chenliang Li and Yuanhong Xu and Hehong Chen and Junfeng Tian and Qian Qi and Ji Zhang and Fei Huang},
   isbn = {2304.14178v1},
   month = {4},
   title = {mPLUG-Owl: Modularization Empowers Large Language Models with Multimodality},
   url = {https://arxiv.org/abs/2304.14178v1},
   year = {2023},
}

@article{Peng2023,
   author = {Zhiliang Peng and Wenhui Wang and Li Dong and Yaru Hao and Shaohan Huang and Shuming Ma and Furu Wei},
   month = {6},
   title = {Kosmos-2: Grounding Multimodal Large Language Models to the World},
   url = {https://arxiv.org/abs/2306.14824v3},
   year = {2023},
}

@misc{zhang2023videollama,
      title={Video-LLaMA: An Instruction-tuned Audio-Visual Language Model for Video Understanding}, 
      author={Hang Zhang and Xin Li and Lidong Bing},
      year={2023},
      eprint={2306.02858},
      archivePrefix={arXiv},
      primaryClass={cs.CL}
}

@misc{maaz2023videochatgpt,
      title={Video-ChatGPT: Towards Detailed Video Understanding via Large Vision and Language Models}, 
      author={Muhammad Maaz and Hanoona Rasheed and Salman Khan and Fahad Shahbaz Khan},
      year={2023},
      eprint={2306.05424},
      archivePrefix={arXiv},
      primaryClass={cs.CV}
}

\end{document}